\documentclass[letterpaper]{article} % DO NOT CHANGE THIS
\usepackage{aaai2027}
\usepackage[hyphens]{url}  % DO NOT CHANGE THIS
\usepackage{graphicx} % DO NOT CHANGE THIS
\usepackage{natbib}  % DO NOT CHANGE THIS AND DO NOT ADD ANY OPTIONS TO IT
\usepackage{caption} % DO NOT CHANGE THIS AND DO NOT ADD ANY OPTIONS TO IT
\usepackage{algorithm}
\usepackage{newfloat}
\usepackage{listings}
\DeclareCaptionStyle{ruled}{labelfont=normalfont,labelsep=colon,strut=off} % DO NOT CHANGE THIS
\floatstyle{ruled}
\newfloat{listing}{tb}{lst}{}
\floatname{listing}{Listing}

\usepackage{multirow}
\usepackage{booktabs}

\usepackage{todonotes}
\usepackage[most]{tcolorbox}
\usepackage{cancel}
\usepackage{amsmath}
\usepackage{amssymb}
\usepackage{algpseudocode}
\usepackage{listings}
\newcommand{\cc}{\textbf{C-PTQ}}
\definecolor{best}{HTML}{D7191C} 
\definecolor{second}{HTML}{008080}

\definecolor{beige}{RGB}{250,240,210}
\definecolor{peach}{RGB}{255,220,180}
\definecolor{deeppeach}{RGB}{230,150,100}
\definecolor{lightpink}{RGB}{255, 235, 240}
\definecolor{mint}{RGB}{213, 238, 216}
\definecolor{lightpurple}{RGB}{235,210,255} 
\definecolor{purple}{RGB}{145,115,180}
\definecolor{softblue}{RGB}{0, 160, 220}
\definecolor{softgreen}{RGB}{90,150,125}
\definecolor{deepgreen}{RGB}{0,128,0} 
\definecolor{midgray}{gray}{0.8}
\definecolor{lightgray}{gray}{0.9}
\definecolor{darkred}{RGB}{178,34,34}
\usepackage{xcolor}
\usepackage[table]{xcolor}

\newcommand{\vv}[1]{\textsubscript{{\texttt{#1}}}}

\newcommand{\sota}[1]{{\textbf{#1}}}

\usepackage{arydshln}

\title{\cc: Fisher-weighted Channel-wise Sensitivity for Post-training \\Quantization of MLLMs}

\author{
    Jiameng Li\corresponding$^1$, Han Zhou$^2$, Matthew B. Blaschko$^1$
}
\affiliations{
    $^1$KU Leuven, $^2$Tiangong University
}

\begin{document}

\maketitle

\begin{abstract}

Multimodal large language models (MLLMs) require huge memory and computational costs, which limits their practical deployment. Post-training quantization (PTQ) techniques offer an efficient solution for model compression and inference acceleration. Yet, the quantized model faces performance degradation due to outlier channels, which are highly sensitive to quantization and substantially impair activation fidelity and task accuracy.
To protect these salient channels during quantization, existing PTQ methods leverage modality- or token-level metrics to guide channel-wise scaling (CWS) of LLM decoders. 
However, these orthogonal measurements fail to capture channel-wise impacts on task-specific loss, and the misalignment between importance and scaling factors ultimately leads to suboptimal performance.
To address this issue, we propose \cc, a unified channel-wise PTQ method that harmonizes task-specific loss perturbation and quantization error. Motivated by second-order derivatives, we design a Fisher-weighted objective as a tractable Hessian approximation, seamlessly injecting task sensitivity into the scaling process. Notably, we achieve state-of-the-art performance without auxiliary modules like LoRA, thereby maintaining high efficiency.
Experiments on Qwen2.5VL, InternVL2 and LLaVA-OV across 8 benchmarks demonstrate our effectiveness in both weight-only and weight-activation settings. Our GitHub repository: {\url{https://github.com/renaissanceee/cptq}}.

\end{abstract}


\section{Introduction}
\label{sec:intro}

\begin{figure}[t]
\centering
\includegraphics[width=\linewidth]
{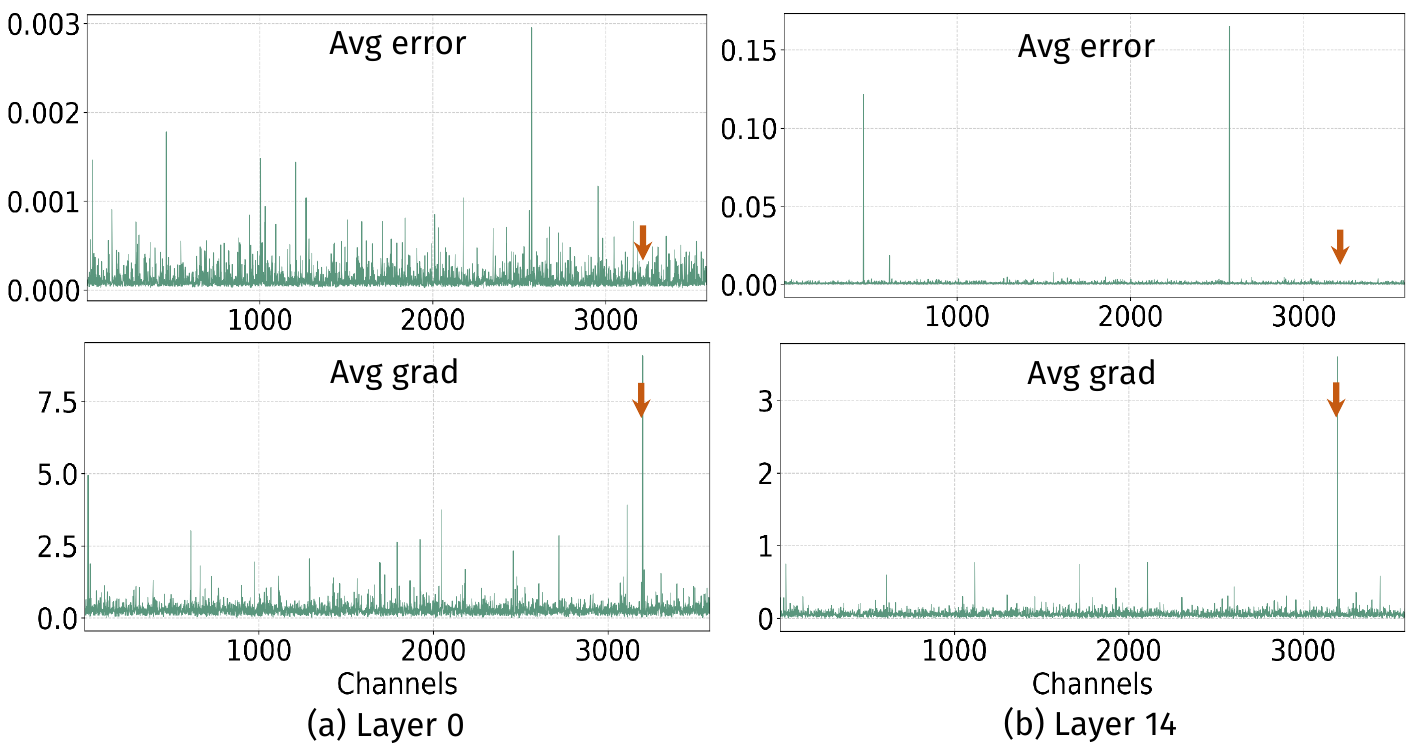} 
\caption{
\textbf{Visualization of channel-wise quantization error and gradients (Qwen2.5VL-7B).} The round-to-nearest squared quantization error and squared gradients show a misalignment (red arrow).
, which compromises the channel-wise scaling search.  
}
\label{fig:vis_ch_qwen}
\end{figure}

The capacity of advanced MLLMs comes with huge computational costs, which limit their practical application. 
Much effort has been dedicated to reducing the computational overheads for developing efficient MLLMs, including distillation \citep{shu2025llava,xu2024llavadi}, pruning \citep{li2026mi,fan2025visipruner} and quantization \citep{li2025mbq,xiang2026qig}. Among them, \textbf{distillation} requires sufficient training data to achieve effective knowledge transfer \citep{lee2026masking}, and \textbf{pruning} 
exhibits limited throughput gains relative to the pruning rate
due to irregular memory access.
In contrast, post-training \textbf{quantization} is 
hardware-friendly and data-efficient, requiring only a small calibration set to achieve rapid deployment on standard hardware.
To mitigate quantization error, channel-wise scaling \citep{lin2023awq} has emerged as a general technique.
However, preserving the task accuracy of MLLMs under low-bit quantization remains challenging.

\begin{figure*}[h]
\centering
\includegraphics[width=\linewidth]
{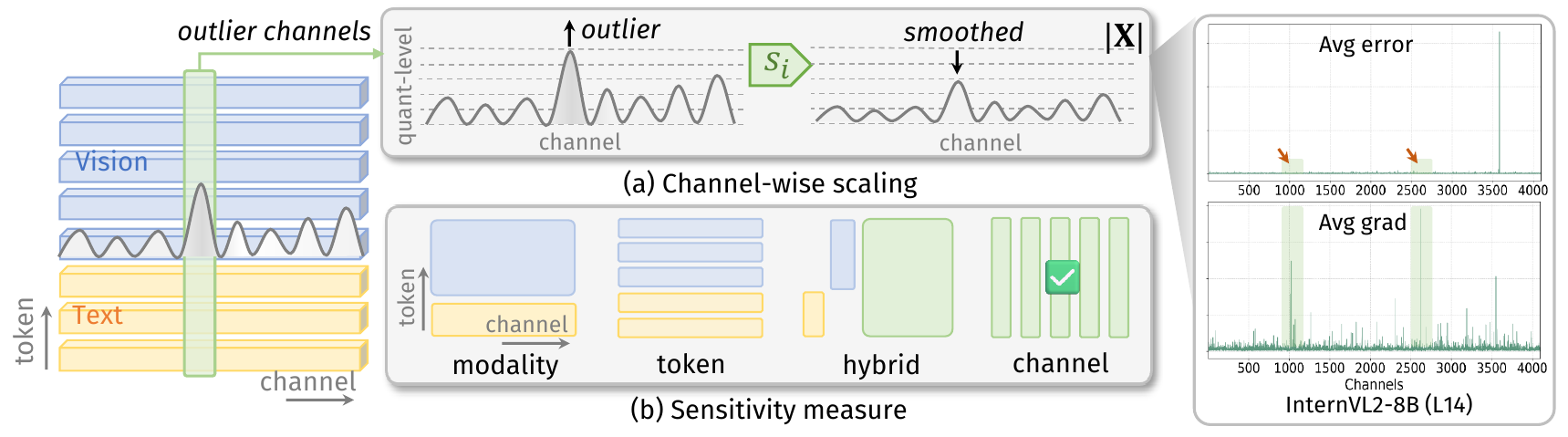} 
%%% The dashed line represents quantization levels.
\caption{
\textbf{Overview.} (a) Channel-wise scaling \citep{lin2023awq} aims to smooth outlier channels with large magnitudes to reduce quantization error. However, channels with minimal errors can still exhibit substantial SFT gradients with degraded performance, which calls for fine-grained sensitivity measures. (b) The block layouts show the interest of dimension on sensitivity metrics, including modality- \citep{li2025mbq}, token- \citep{xiang2026qig,zhouimpquant}, hybrid- \citep{jia2026quant,zhong2026breaking} and channel-wise measures (ours). We unify the channel-wise sensitivity and scaling for effective quantization.
}
\label{fig:overview}
\end{figure*}

%%%%%%%%%%%%%%%%%%%%%%%%%%%%%
As shown in Fig.~\ref{fig:vis_ch_qwen}, there is a misalignment between quantization error and task-specific loss.
Specifically, the channel around the $3200$-th with small quantization error corresponds to large SFT gradients, instead of nearly zero. As a result, the optimization towards minimizing quantization error leads to suboptimal task performance.
This motivates more fine-grained sensitivity measures beyond activation magnitudes to align with SFT gradients. 
For instance, MBQ \citep{li2025mbq} and QIG \citep{xiang2026qig} extend the quantization objective to be modality-aware and token-aware. The first work MBQ holds the best efficiency compared to the following work \citep{xiang2026qig,hu2026masquant}, but the performance depends on the degree of modality imbalance in pretrained MLLMs. 
Recent work \citep{zhong2026breaking,jia2026quant} introduces low-rank matrices with hybrid importance measures to compensate for quantization error, which also brings more computation overhead. 
Interestingly, although the scaling factor is searched per channel (Fig.~\ref{fig:overview}(a)), the quantization sensitivity was investigated from the view of modality-, token-, or hybrid-heterogeneity (Fig.~\ref{fig:overview}(b)).
This leads to a natural question: \textit{Can we directly inject channel-wise importance into CWS for aligned importance and scaling search?}

To support the synergy, 
%%%%%%%%%%%%%%%%%%%%%%%%
we propose \cc, a unified PTQ method that explicitly aligns channel-wise scaling with global task sensitivity. Our approach is built on the premise that minimizing the quantization error is insufficient; \textbf{an effective strategy must reflect the distinct impact of individual channels on task sensitivity}. To bridge this gap, we leverage second-order curvature information. Specifically, we introduce a Fisher-weighted quantization objective as a tractable approximation of the Hessian matrix. By applying a diagonal simplification~\citep{lecun1989optimal}, we transform the complex loss perturbation into a decoupled objective of channel-wise weighted quantization error. Consequently, \cc~directly penalizes the reconstruction residual of each channel by its gradient variance. By integrating activation statistics and gradient sensitivity into the channel axis, our method achieves state-of-the-art performance while avoiding computationally intensive operations.

Our key contributions are summarized as follows:
\begin{itemize}
    \item We propose \cc, a unified post-training quantization method that explicitly aligns channel-wise importance with scaling factors for fine-grained quantization.    
    \item By employing a diagonal Fisher matrix as a tractable Hessian approximation, we derive an efficient Fisher-weighted objective that directly embeds task-specific sensitivity into the quantization process, thereby preserving downstream performance (Sec.~\ref{sec:methods}).

    \item Extensive experiments on Qwen2.5VL, InternVL2 and LLaVA-OV show that our method achieves SOTA performance with favorable efficiency (Sec.~\ref{sec:exp}).
\end{itemize}

%%%%%%%%%%%%%%%%%%%%%%%%%%%%%%%
\section{Related Work}
\label{sec:related}

\paragraph{PTQ for MLLMs.}   
Despite the practical efficiency, quantization fundamentally struggles with large-magnitude outliers concentrated in specific channels, which stretch the quantization range and undermine low-bit representations. 
Recognizing the critical role of activations in determining quantization error,
early \textbf{activation-aware} PTQ methods like AWQ \citep{lin2023awq} and SQ \citep{xiao2023smoothquant} introduce channel-wise scaling by the activation distributions.
%%%%%%%%%%%%%%%%%%%%%%%
To capture structural importance beyond activation magnitudes, \textbf{curvature-based} methods like GPTQ \citep{frantar2023gptq} incorporate second-order Hessian to apply closed-form error compensation during quantization. Inspired by fine-tuning techniques, \textbf{LoRA-based} methods \citep{hu2026masquant,jia2026quant} seek to push performance boundaries by applying external low-rank matrices for fine-grained compensation. 
Yet, LoRA itself incurs additional optimization and inference costs, motivating an efficient solution that directly aligns channel-wise scaling with task sensitivity.

\paragraph{Sensitivity measure.}
Compared with LLMs, the PTQ of MLLMs necessitates elaborate sensitivity measurements for the intertwined multimodal nature.
Motivated by the issue of \textbf{modality imbalance}, MBQ \citep{li2025mbq} reweights the quantization error of AWQ \citep{lin2023awq} using the average vision and text gradients. Building upon this idea, MABA \citep{zhang2026maba} employs modality-specific gradients for mixed-precision bit allocation, while MASQuant \citep{hu2026masquant} enhances SQ \citep{xiao2023smoothquant} with modality-aware smoothing factors. As an extension of GPTQ \citep{frantar2023gptq}, VLM-PTQ \citep{deng2026vlm} handles vision and text Hessians separately before fusing them for quantization.
Besides, QIG \citep{xiang2026qig} leverages integrated gradients to characterize \textbf{token-wise} sensitivity, enabling the quantization objective to place greater emphasis on critical tokens. The following work considers a \textbf{hybrid} heterogeneity, which compensates for a few token-dependent \citep{jia2026quant} or modality-specific \citep{zhong2026breaking} channels with low-rank matrices. 
Despite the progress, we observe a fundamental mismatch: sensitivity is typically measured at the modality or token level, whereas scaling is optimized at the channel level.
To address this discrepancy, we revisit \textbf{Fisher Information}, a classical sensitivity metric widely used for model compression \citep{cherilyn2026supernodes}
and parameter-efficient adaptation \citep{sung2021training}. Please refer to App.~\ref{sec:app_related} for more illustration.
By incorporating a Fisher-weighted objective into channel-wise scaling, we build a synergy between the activation statistics and the curvature information.

%%%%%%%%%%%%%%%%%%%%%%%%
\section{Methods}
\label{sec:methods}
%%%%%%%%%%%%%%%%%%%%%%%%

\subsection{Preliminaries}
\label{subsec:method_pre}
Let $\mathbf{Y}{=}\mathbf{WX} \in \mathbb{R}^{N_c^\mathrm{out}\times N_k}$ denote the full-precision output of a linear layer, where $N_c^\mathrm{out}$ and $N_k$ represent the number of output channels and tokens, respectively. Throughout this paper, we use {{WxAy}} to indicate $x$-bit weight and $y$-bit activation quantization. Taking weight-only quantization as an example, the quantization function $Q$ maps the high-precision $\mathbf{W}$ into discrete levels (\textit{e.g.}, from FP16 to INT4), which introduces a reconstruction residual $\Delta\mathbf{Y}{=}\mathbf{W}\mathbf{X} {-} Q(\mathbf{W})\mathbf{X}$. To protect important channels and mitigate the reconstruction error, channel-wise scaling approaches introduce a scaling vector $\mathbf{s}$, reformulating the residual as:
\begin{align}
\Delta\mathbf{Y}&=\mathbf{W}\mathbf{X} {-} Q(\mathbf{W} * \mathbf{s})(\mathbf{s}^{-1} * \mathbf{X}).
\label{eq:delta_y}
\end{align}
To find the optimal scaling factor $\mathbf{s}$, AWQ \citep{lin2023awq} conducts a grid search for each LLM layer to minimize Mean Squared Error
(MSE) of residuals:
\begin{align}
\mathbf{s}^*&=\arg \min_{\mathbf{s}} \|\Delta\mathbf{Y}\|_2^2.
\label{eq:awq}
\end{align}
When extending channel-wise scaling to MLLMs, existing methods leverage modality- or token-level sensitivity for reweighted quantization objectives \citep{li2025mbq,xiang2026qig} or adaptive bit allocation \citep{zhang2026maba}. For instance, MBQ \citep{li2025mbq} splits the residual $\Delta\mathbf{Y}$ from Eqn.~(\ref{eq:delta_y}) into visual and textual components $\Delta\mathbf{Y}^v $ and $ \Delta\mathbf{Y}^t$, to minimize a modality-balanced quantization error:
\begin{align}
\mathbf{s}^*&=\arg \min_{\mathbf{s}}\sum_i^{N_k^v}|\bar{\mathbf{g}}^v| \cdot\|\Delta\mathbf{y}^v_i\|+\sum_i^{N_k^t}|\bar{\mathbf{g}}^t|\cdot\|\Delta\mathbf{y}^t_i\|.
\label{eq:mbq}
\end{align}
Here, $|\bar{\mathbf{g}}^{v}|$ and $|\bar{\mathbf{g}}^{t}| \in \mathbb{R}_{\geq0}$ denote the average absolute gradients of the $N_k^v$ vision and $N_k^t$ text tokens, respectively. 
%%%%%%%%%%%%%%%%
Taking a step further into token-level granularity, QIG \citep{xiang2026qig} enhances the objective at the token level using token-wise integrated gradients $\lambda_i$:
\begin{align}
\mathbf{s}^*&=\arg \min_{\mathbf{s}} \sum_{i=1}^{N_k}\lambda_i \cdot\|\Delta\mathbf{y}_i\|_2^2,
\label{eq:qig}
\end{align}
where $\Delta\mathbf{y}_i$ denotes the residual of the $i$-th in overall $N_k$ tokens.
%%%%%%%%%%%%%%%%%%%%%%%%%%%%%%%%%
Recent work introduces a hybrid paradigm \citep{jia2026quant,zhong2026breaking}, which splits a few outlier channels for separate compensation. 
A closer look at these approaches reveals a fundamental misalignment: although sensitivity is evaluated across the modality or token dimension, the optimization is executed per-channel, as shown in Fig.~\ref{fig:overview}. This discrepancy leads to a natural question: \textit{Can we directly align channel-wise importance with channel-wise scaling?}

\subsection{Fisher-weighted Quantization}
\label{subsec:method_Fisher}

\paragraph{Loss perturbation.}
To explicitly align channel-wise quantization error with task-specific sensitivity, we bridge the output residual $\Delta\mathbf{Y}$ with the overall loss perturbation $\Delta\mathcal{L}$. Let $\hat{\mathbf{Y}}$ denote the quantized output of the original $\mathbf{Y}$, written as $\hat{\mathbf{Y}} = \mathbf{Y}+\Delta\mathbf{Y}$. Assuming $\mathcal{L}(\mathbf{Y})$ is the SFT loss (\textit{e.g.} Cross-Entropy), we first analyze the second-order Taylor expansion of $\mathcal{L}$ around the full-precision output $\mathbf{Y}$: 
\begin{equation}
    % \mathcal{L}(\mathbf{Y} + \Delta\mathbf{Y}) 
    \mathcal{L}(\mathbf{\hat{Y}}) 
    {=} \mathcal{L}(\mathbf{Y}) 
    + \underbrace{\mathbf{g}^{\top} \Delta\mathbf{Y}}_{\text{first-order}} 
    + \underbrace{\frac{1}{2}\,\Delta\mathbf{Y}^{\top} \mathbf{H}\, \Delta\mathbf{Y}}_{\text{second-order}} 
    + \cancel{\mathcal{O}(\|\Delta\mathbf{Y}\|^3)},
\end{equation}
where $\mathbf{g} = \frac{\partial\mathcal{L}}{\partial\mathbf{Y}}$ and 
$\mathbf{H} = \frac{\partial^2\mathcal{L}}{\partial\mathbf{Y}^2}$ are the gradient and Hessian matrices of the output activation $\mathbf{Y}$. Dropping the third-order remainder \citep{lecun1989optimal,frantar2023gptq}, the task loss perturbation ($\Delta\mathcal{L}$) caused by quantization can be approximated as:
\begin{equation}
    \Delta\mathcal{L} \;=\; \mathcal{L}(\hat{\mathbf{Y}}) - \mathcal{L}(\mathbf{Y}) 
    \;\approx\; \mathbf{g}^{\top}\Delta\mathbf{Y} + \frac{1}{2}\,\Delta\mathbf{Y}^{\top}\mathbf{H}\,\Delta\mathbf{Y}.
    \label{eq:delta_l}
\end{equation}

\paragraph{Vanishing first-order terms.}

Following the Optimal Brain Damage (OBD) framework \citep{lecun1989optimal}, we assume that the pretrained model has already converged to a local optimum. Consequently, the expected gradient evaluated on the calibration set is close to zero, with the negligible first-order term. Thus, the performance degradation is predominantly governed by the second-order curvature term, aligning with prior studies \citep{frantar2023gptq,lecun1989optimal}:
\begin{equation}
    \Delta\mathcal{L} \;\approx\; \frac{1}{2}\,\Delta\mathbf{Y}^{\top}\mathbf{H}\,\Delta\mathbf{Y}.
    \label{eq:second_order}
\end{equation}
Crucially, the quadratic form in Eqn.~(\ref{eq:second_order}) reveals that the loss perturbation ($\Delta\mathcal{L}$) is not strictly equivalent to the uniform quantization error ($\|\Delta\mathbf{Y}\|^2$), but is modulated by the Hessian matrix. This theoretical misalignment accounts for the misplacement in Fig.~\ref{fig:vis_ch_qwen}, which motivates our channel-wise PTQ approach. The extensive empirical analysis retaining the first-order term is provided in App.~\ref{subsec:app_ablation}.
% Sec.~\ref{subsec:exp_ablation}.

\begin{figure}[!hb]
\centering
\includegraphics[width=\linewidth]
{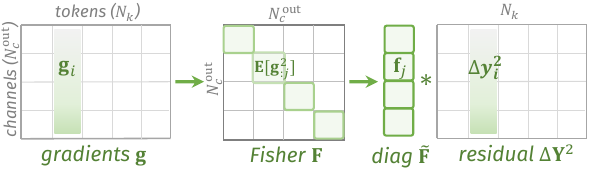} 
\caption{
\textbf{Our Fisher-weighted quantization objective.}
The diagonal Fisher $\mathbf{\tilde{F}}$ works as channel-wise importance in quantization objective. For simplification, we assume tokens $N_k{=}5$ and output channels $N_c^\mathrm{out}{=}4$ here.
}
\label{fig:Fisher}
\end{figure}

\paragraph{From Hessian to Fisher Information.}
Since computing the exact Hessian matrix is 
% computationally prohibitive 
less feasible
for MLLMs, we approximate $\mathbf{H}$ using the empirical Fisher Information Matrix $\mathbf{F}$ computed on the calibration set.
Denote $\mathbf{g}_i \in \mathbb{R}^{N_c^\mathrm{out}}$ as the vectorized output gradient for token $i$, the empirical Fisher is written as the expectation of the outer products of gradients over all tokens.
\begin{align}
    \mathbf{H} &\approx \mathbf{F} = \mathbb{E}\left[\mathbf{g}_i\mathbf{g}_i^{\top}\right].
    \label{eq:h_f}
\end{align}
Substituting this into Eqn.~(\ref{eq:second_order}), the loss perturbation is reformulated as:
\begin{align}
    \Delta\mathcal{L} &\approx 
    \frac{1}{2}\,\Delta\mathbf{Y}^{\top}\,\mathbb{E} \left[\mathbf{g}_i\mathbf{g}_i^{\top}\right]\Delta\mathbf{Y}.
    \label{eq:delta_l_a}
\end{align}

\noindent Unlike GPTQ \citep{frantar2023gptq}, which derives an activation-based Hessian ($\mathbf{H}=\mathbf{X}\mathbf{X}^\top$) to minimize the local reconstruction error, our Fisher matrix intrinsically encapsulates the global task loss sensitivity. Further empirical comparisons are detailed in Sec.~\ref{subsec:exp_ablation}.

\paragraph{Diagonal approximation.}
To integrate this curvature information with channel-wise scaling, we decouple the channel dimensions and adopt a diagonal approximation of the Fisher matrix.
%%%%%%%%%%%%%%%%%%%%%%%%%%%%%
Let $i,j$ be the token and channel index and $\mathbf{g}_{ij}$ denote the gradient with respect to the output $\mathbf{y}_{ij}$. Assuming $\mathbf{f}$ represents the diagonal elements of $\mathbf{F}$, the diagonalized matrix $\mathbf{\tilde{F}}$ assigns a uniform Fisher weight $\mathbf{f}_j$ to all tokens within channel $j$:
\begin{equation}
    \mathbf{F} \approx \mathbf{\tilde{F}} =\operatorname{diag}(\mathbf{f}), 
    \text{where}~ \mathbf{f}_j = \mathbb{E}_{:,j}\left[\mathbf{g}_{ij}^2\right].
    \label{eq:Fisher_diag}
\end{equation}
%%%%%%%%%
Thereby, Eqn.~(\ref{eq:delta_l_a}) can be simplified into a channel-wise weighted residual:
\begin{align}
    \Delta\mathcal{L}\ &\approx
   \frac{1}{2}\Delta\mathbf{Y}^{\top}\mathrm{diag}(\mathbf{f})\,\Delta\mathbf{Y}
    =\frac{1}{2} \sum_{i=1}^{N_k}\|\mathbf{\sqrt{f}} \odot
\Delta\mathbf{y}_i \|_2^2.
    \label{eq:delta_l_c}
\end{align}
Eqn.~(\ref{eq:delta_l_c}) translates the loss perturbation into a simple, element-wise weighted residual, with which we successfully bridge the optimization gap between task-specific loss and quantization error. This explicit alignment guarantees that \cc~can efficiently execute channel-wise scaling while being strictly guided by task sensitivity.

\subsection{Unified Channel-wise Scaling}
\label{subsec:method_ch}

\paragraph{Optimization objective.}
By substituting the scaled residual $\Delta\mathbf{Y}$ in Eqn.~(\ref{eq:delta_y}) into Eqn.~(\ref{eq:delta_l_c}), our ultimate goal is to find the optimal scaling vector $\mathbf{s}^*$ that minimizes the Fisher-weighted quantization error $\mathrm{Err}(\mathbf{s})$:
\begin{align}
    \min_{\mathbf{s}}\|\mathbf{\sqrt{f}} * \left(\mathbf{W}\mathbf{X} - (\mathbf{W} * \mathbf{s})\,Q(\mathbf{s}^{-1} * \mathbf{X})\right)\|_2^2,
    \label{eq:err_a}
\end{align}
where * denotes channel-wise multiplication. 
When extending our framework to weight-activation quantization, the objective becomes:
\begin{align}
    \min_{\mathbf{s}}\|\mathbf{\sqrt{f}} * \left(\mathbf{W}\mathbf{X} - Q(\mathbf{W} * \mathbf{s})\,Q(\mathbf{s}^{-1} * \mathbf{X})\right)\|_2^2.
\label{eq:err_b}
\end{align}
On the calibration set $\mathcal{D}_{\mathrm{cal}}$, the per-channel Fisher weights $\mathbf{f}_j$ are estimated by the average squared gradients, and subsequently normalized by their mean across all channels to stabilize the magnitude.
Therefore, Eqn.~(\ref{eq:err_a}) or (\ref{eq:err_b})  is implemented as:
\begin{align}
\text{Err}(\mathbf{s}) \propto \sum_{i=1}^{N_k} \sum_{j=1}^{N_c^\mathrm{out}} \mathbf{f}_j\cdot \Delta\mathbf{y}_{ij}^2.
 \label{eq:err_c}
\end{align}

\begin{table*}[!h]
\centering
\caption{\textbf{Performance on Qwen2.5VL.} We report results on 7B and 32B models. The weight-only setting W3A16 quantizes weights into 3-bit, and weight-activation settings W4A8 and W4A6 quantize weights into 4-bit and activations into 8-bit and 6-bit, respectively. 
We use \sota{bold} to indicate SOTA results.
} 
\label{tab:results_qwen2.5}
\resizebox{0.99\textwidth}{!}{
\begin{tabular}{llcccccccc|cc}
\toprule
\textbf{Setting} & \textbf{Method}
& \textbf{MMMU}  & \textbf{SQA} & \textbf{VizWiz} & \textbf{TextVQA} & \textbf{OCRBench} & \textbf{ChartQA}  & \textbf{AI2D} & \textbf{DocVQA} & \textbf{Avg} & \textbf{\%} \\

\hline
\rowcolor{lightgray}
FP16 & Qwen2.5VL-7B 
% \multicolumn{2}{l}{Qwen2.5VL-7B}
& 50.8 & 87.2 & 70.5 & 83.1 & 83.9 & 83.1 & 82.9 & 95.7  & 79.7 & 100 \\
\hline
\multirow{5}{*}{\textit{W3A16}} & AWQ\vv{MLSys'24}   
& 46.7 & 84.9 & 65.0 & 80.9 & 81.0 & 80.4 & 79.9 & 94.3 & 76.6 & -3.9 \\
& MBQ\vv{CVPR'25} 
& 46.8 & 85.2 & 66.2 & \sota{81.0} & 81.6 & \sota{82.0} & 79.3 & 94.0 & 77.0 & -3.4 \\
& QIG\vv{CVPR'26} 
& 44.2 & 84.0  & \sota{66.9} & 80.8 & 81.7 & 81.4  & \sota{80.2} & 94.3 & 76.7 & -3.8 \\
& MASQuant\vv{CVPR'26} 
& 41.9 & 81.0 & 65.4 & 74.3 & 57.9 & 77.1 & 75.6 & 92.0 & 70.7 & -11.3 \\
& \cc 
& \sota{48.4} & \sota{85.2} & 64.6 & 80.2 & \sota{82.6} & 80.9 & 79.9 & \sota{94.9} & \sota{77.1} & \sota{-3.3} \\
\hline
\multirow{5}{*}{\textit{W4A8}} & SQ\vv{ICML'23}
& 41.6 & 73.0 & 65.5 & \sota{77.3} &  77.0 & 79.9 & 73.4 & 91.1 & 72.4 & -9.2 \\
& MBQ\vv{CVPR'25}
& \sota{45.0}  & 84.1 & 59.2 &  71.8 & 72.3 & 73.6 & 79.4 & 86.2 & 71.5 & -10.3 \\
& QIG\vv{CVPR'26}
& 41.9 & 84.4 & 63.6  & 76.0 & \sota{78.2} & 76.7 & 79.4 & \sota{90.7} & 73.9 & -7.3 \\
& MASQuant\vv{CVPR'26} 
& 43.3 & \sota{85.7} & \sota{66.4} & 77.0 & 72.8 & \sota{82.8} & \sota{80.6} & 87.4 & 74.5 & -6.5 \\

& \cc 
& 44.0 & 85.0 & 64.8 & {77.1} & 77.2 & 81.6 & 78.9 & 90.0 & \sota{74.8} & \sota{-6.1} \\
\hline
\multirow{5}{*}{\textit{W4A6}}  & SQ\vv{ICML'23}
& 29.0 & 75.0 & 63.9 & 74.6 & 76.1 & 73.2 & 70.3 & 92.4 & 69.3 & -13.3 \\
& MBQ\vv{CVPR'25}
& 38.4 & 73.0 & 48.2 & 65.2 & 65.9 & 58.8 & 70.6 & 84.5 & 63.1 & -20.8 \\
& QIG\vv{CVPR'26}
& 39.1 & 71.5 & 56.9 & 70.9 & 74.3 & 67.4 & 71.5 & 89.8 & 67.7 & -15.1 \\
& MASQuant\vv{CVPR'26}
& 29.7 & 79.7 & 62.6 & 72.9 & 70.3 & \sota{73.3} & \sota{78.3} & 85.1 & 69.0 & -13.4 \\

& \cc 
& \sota{41.8} & \sota{81.6} & \sota{62.8} & \sota{76.2} & \sota{78.3} & 70.6 & 75.7 & \sota{92.4} & \sota{72.4} & \sota{-9.2} \\

\hline

\rowcolor{lightgray}
% Qwen2.5VL-32B & -
% \multicolumn{2}{l}{Qwen2.5VL-32B}
FP16 & Qwen2.5VL-32B 
& 60.2 & 91.6 & 64.7 & 77.3 & 79.8 & 69.2 & 84.5  & 94.6 & 77.7 & 100 \\
\hline
\multirow{4}{*}{\textit{W3A16}} & AWQ\vv{MLSys'24}   
& 55.7 & 89.2 & 63.0 & 77.8 & 77.6 & \sota{69.4} & 82.8 & 95.4 & 76.4 & -1.7 \\
& MBQ\vv{CVPR'25}
& \sota{58.8} & 87.4 & 62.4 & 78.0 & 77.0 & 73.4 & 82.5 & 95.0 & 76.6 & -1.4 \\
& QIG\vv{CVPR'26} 
& 55.6  & 88.2 & 62.3 & 77.1 & \sota{78.5} & 65.6 & 82.8 & 95.2 & 75.7 & -2.6 \\
& \cc 
& 57.3 & \sota{89.8} & \sota{64.6} & \sota{78.4} & 78.2 & 67.8 & \sota{83.9} & \sota{96.3} & \sota{77.0} & \sota{-0.9} \\

\hline
\multirow{4}{*}{\textit{W4A8}} &  SQ\vv{ICML'23}
& 56.2 & 87.4 & \sota{66.3} & 75.9 & 75.6 & \sota{69.6} & 81.9 & 93.5 & 75.8 & -2.4 \\
& MBQ\vv{CVPR'25} 
&  54.4 & 87.5 & 57.3 & 73.7 & 75.4 & 67.5 & 81.1 & 92.0 & 73.6 & -5.3 \\
& QIG\vv{CVPR'26}
& 54.9 & \sota{87.7} & 64.2 & 76.1 & 78.1 & \sota{69.6} & 82.2 & 94.5  &  75.9 & -2.3 \\

& \cc
& \sota{58.0} & 87.5 & 65.1 & \sota{76.3} & \sota{79.1} & 66.2 & \sota{82.4} & \sota{94.6} & \sota{76.2} & \sota{-1.9} \\

\hline
\multirow{4}{*}{\textit{W4A6}} & SQ\vv{ICML'23} 
& 51.1 & 85.1 & 58.2 & \sota{73.4} & 75.2 & \sota{61.5} & 79.5 & 92.9 & 72.1 & -7.2 \\
& MBQ\vv{CVPR'25}
& 49.7 & 82.1 & 54.0 & 68.5 & 72.0 & 49.6 & 80.0 & 86.8 & 67.8 & -12.7 \\
& QIG\vv{CVPR'26}
& 49.1 & 84.3 & \sota{61.4} & 72.4 & 75.2 & 54.0 & 78.7 & \sota{93.0} & 71.0 & -8.6 \\
& \cc 
& \sota{55.8} & \sota{86.0} & 59.1 & 71.5 & \sota{77.6} & 57.4 & \sota{81.2} & 92.7 & \sota{72.7} & \sota{-6.4} \\
\hline
\end{tabular}
}
\end{table*}

As illustrated in Fig.~\ref{fig:Fisher}, this objective selectively penalizes channels that are highly critical to the task loss $\mathcal{L}$, overcoming the limitation of the vanilla MSE in AWQ while remaining activation-aware. 
Unlike MBQ and QIG, our reweighting and scaling directions are aligned along the channel axis. Although a dense Fisher matrix theoretically accounts for cross-channel dependencies, our empirical results indicate that our diagonal simplification performs on par with the dense one (see details in Sec.~\ref{subsec:exp_ablation}). Given the high computational cost of dense matrices, the proposed diagonalized approximation provides a practical and efficient alternative. The overall pipeline of \cc~is presented in Alg.~\ref{alg:ours}.

%%%%%%%%%%%%%%%%%%%%%%%%%%%%%%%%%%%%
\begin{algorithm}[!h]
\caption{\cc
~with Fisher-weighted MSE
}
\label{alg:ours}
\begin{algorithmic}[1]
\State \textbf{Input:} The gradient matrix $\mathbf{g}$ of output $\mathbf{Y}{\in} \mathbb{R}^{N_c^\mathrm{out}\times N_k}$, weight $\mathbf{W} {\in} \mathbb{R}^{N_c^\mathrm{out} \times N_c^\mathrm{in}}$, input activation $\mathbf{X} {\in} \mathbb{R}^{N_c^\mathrm{in} \times N_k}$, quantization function $Q$, calibration set $\mathcal{D}_\mathrm{cal}$.

\State \textbf{Objective:} 
$\arg 
\min_{\mathbf{s}^*} L(\mathbf{\hat{Y}}(\mathbf{s}))-L(\mathbf{Y})$

\State \textbf{Output:}
Scaling matrix $\mathbf{s}{\in} \mathbb{R}^{N_c}$.

%%%%%%%%%%%%%%%%%%%%%%%%%%%%%%%%%%%%
\State
% \textcolor{softgreen}
{\textit{// Step 1. Channel-wise sensitivity (diagonal Fisher)}}

\State $\mathbf{F}\leftarrow\mathbb{E}[\mathbf{g}_i\mathbf{g}_i^\top],\mathbf{f}_j\leftarrow\mathbb{E}_{:,j}[\mathbf{g}_{ij}^2]$
\State $\{\mathbf{f}_0,...,\mathbf{f}_{N_c}\} \leftarrow \mathrm{mean}_{:,j}(\mathbf{g}_{ij}^2)$

\State
% \textcolor{softgreen}
{\textit{// Step 2. Channel-wise scales (activation-aware)}}
\State $\{\mathbf{x}_0,...,\mathbf{x}_{N_c}\} \leftarrow\mathrm{mean}_{:,j}(\|\mathbf{X}_{ij}\|)$ 
\State $\mathbf{s}\leftarrow \{\mathbf{x}_0^\alpha,...,\mathbf{x}_{N_c}^{\alpha}\}$

\State
% \textcolor{softgreen}
{\textit{// Step 3. Grid search with Fisher-weighted MSE over $\mathcal{D}_\mathrm{cal}$ (*: channel-wise multiplication)}}
\State $\Delta\mathbf{Y}\leftarrow\mathbf{WX}-Q(\mathbf{W}*\mathbf{s})(\mathbf{s}^{-1}*\mathbf{X})$\\
or $\Delta\mathbf{Y}\leftarrow\mathbf{WX}-Q(\mathbf{W}*\mathbf{s})Q(\mathbf{s}^{-1}*\mathbf{X})$  
\State $\mathrm{Err}(\mathbf{s})\leftarrow  \| \mathbf{\sqrt{f}} *\Delta\mathbf{Y}\|_2^2, \mathbf{s}^*\leftarrow\arg\min_{\mathcal{D}_\mathrm{cal}}\mathrm{Err}(\mathbf{s})$ 

\State \Return $\mathbf{s}$.

\end{algorithmic}
\end{algorithm}

%%%%%%%%%%%%%%%%%%%%%%%%
\section{Experiments}
\label{sec:exp}
%%%%%%%%%%%%%%%%%%%%%%%%

\subsection{Setup}
\label{subsec:exp_setup}

\paragraph{Models and baselines.} 
To evaluate our approach, we apply quantization to the LLM decoders of three representative pretrained MLLMs: Qwen2.5-VL \citep{qwen2.5-VL}, InternVL2 \citep{chen2024far}, and LLaVA-OV \citep{li2024llavaov}. These models are selected to cover varying degrees of modality bias and activation characteristics. 
For a fair comparison, all channel-wise scaling methods, including AWQ \citep{lin2023awq}, MBQ \citep{li2025mbq}, QIG \citep{xiang2026qig} and our \cc~conduct a 40-step grid search to identify the optimal scaling factors. For LoRA-based MASQuant \citep{hu2026masquant}, we train it for 2 epochs following the default configuration. All baselines are implemented using their official repositories, and the final evaluations are conducted on the LMMs-Eval framework \citep{zhang2025lmms}. Please refer to App.~\ref{subsec:app_imple} for implementation details.

\paragraph{Evaluation.} 
%%%%%%%%%%%%%%%%%%%%%
For the quantization process, we randomly sample 128 image-caption pairs from COCO caption dataset \citep{lin2014microsoft} proposed by
ShareGPT4V \citep{chen2024sharegpt4v}, following previous work \cite{li2025mbq,xiang2026qig}. 
%%%%%%%%%%%%%%%%%%%%%
To ensure a fair comparison, all baselines are calibrated on this same subset. Once calibrated, we evaluate the models on a comprehensive suite of benchmarks: MMMU \citep{yue2023mmmu} for visual reasoning, VizWiz \citep{gurari2018vizwiz} for visual perception, SQA-IMG \citep{lu2022learn}, OCRBench \citep{liu2024ocrbench}, and TextVQA \citep{singh2019towards} for text recognition, ChartQA \citep{masry2022chartqa} and AI2D \citep{kembhavi2016diagram} for structural understanding, and DocVQA \citep{mathew2021docvqa} for document comprehension. Dataset descriptions and our application on LLMs can be found in App.~\ref{subsec:app_dataset} and ~\ref{subsec:app_ablation}.

\subsection{Results}
\label{subsec:exp_results}

\paragraph{Weight-only quantization.}
We first evaluate weight-only quantization under the W3A16 setting. As reported in Tab.~\ref{tab:results_qwen2.5}, the fine-grained sensitivity metrics of QIG \citep{xiang2026qig} and MASQuant \citep{hu2026masquant} do not translate into noticeable average gains, while AWQ \citep{lin2023awq} remains a highly competitive baseline under multimodal calibration data. Against these competitive baselines, \cc~still achieves consistent improvements. Notably, it is the only method that surpasses AWQ on InternVL2-8B (Tab.~\ref{tab:results_intern2_llava}), while also maintaining the top performance on LLaVA-OV-7B. 
In general, the performance gap between weight-only to the full-precision model is relatively small, with all PTQ methods averaging merely a 2-3\% drop on InternVL2-8B. We subsequently explore weight-activation quantization, where the performance degradation becomes more severe.

\begin{table*}[!h]
\centering
\caption{\textbf{Performance on InternVL2 and LLaVA-OV.} We report results on InternVL2-8B and LLaVA-OV-7B, where \cc~demonstrates SOTA performance consistently. 
% Best values are in \textcolor{best}{red}, and second-best in \textcolor{second}{teal}.
}
\label{tab:results_intern2_llava}
\resizebox{0.99\textwidth}{!}{
\begin{tabular}{llcccccccc|cc}
\toprule
% \textbf{Method} 
% \multicolumn{2}{l}{\textbf{Method}}
\textbf{Setting} & \textbf{Method}
& \textbf{MMMU}  & \textbf{SQA} & \textbf{VizWiz} & \textbf{TextVQA} & \textbf{OCRBench} & \textbf{ChartQA}  & \textbf{AI2D} & \textbf{DocVQA} & \textbf{Avg} & \textbf{\%} \\
\hline
\rowcolor{lightgray}
% InternVL2-8B & \textit{FP16}
% \multicolumn{2}{l}{InternVL2-8B}
FP16 & InternVL2-8B
& 47.9 &  97.1 & 61.1 & 77.1 & 76.9 &  82.4 & 82.4 & 91.8 &  77.1 & 100 \\
\hline
\multirow{4}{*}{\textit{W3A16}} & AWQ\vv{MLSys'24} 
& \sota{47.1} & 96.3 & \sota{60.3} & 74.5 & 74.7 & 79.2 & 80.2 & 90.4  & 75.3 & -2.3 \\
& MBQ\vv{CVPR'25}
& 45.2 & 96.3 & 58.5 & 74.9 & 75.3 & 79.9 & 79.5 & 90.0 & 75.0 & -2.7 \\
& QIG\vv{CVPR'26}
& 46.2 & 96.3 & 59.9 & 74.3 & 74.1 & \sota{80.5} & 80.0 & \sota{90.6} & 75.2 & -2.5 \\
& \cc 
& 45.5 & \sota{96.5} & 59.9 & \sota{75.0} & \sota{75.9} & 80.4 & \sota{80.4} & 90.5  & \sota{75.5} & \sota{-2.1} \\

\hline
\multirow{4}{*}{\textit{W4A8}} & SQ\vv{ICML'23}
& 42.2 & 96.0  & 52.1 & 72.3 & 72.0 & 78.3 & 77.1 & 87.8 & 72.2 & -6.4 \\
&  MBQ\vv{CVPR'25}
& 43.9 & 96.1 & 55.9 & 71.9 & 72.9 & 78.2 & 79.4 & 88.8 & 73.4  & -4.8 \\
& QIG\vv{CVPR'26} 
& 43.6 & 95.1 & 56.6 & 72.4  & 72.0 & 77.5 & 77.3 & 88.1  & 72.8 & -5.6 \\

& \cc 
& \sota{45.3} & \sota{96.8} & \sota{57.7} & \sota{72.4} & \sota{73.6} & \sota{78.4} & \sota{80.3} & \sota{89.1} & \sota{74.2} & \sota{-3.8} \\

\hline

\multirow{4}{*}{\textit{W4A6}} & SQ\vv{ICML'23}
& \sota{43.8} &  \sota{95.0} & 51.1 & 69.9 & 70.6 & \sota{75.3} & 74.8 & 86.7 & 70.9 & -8.0 \\
& MBQ\vv{CVPR'25} 
& 41.4 & 94.7 & 53.3 & 68.9 & 69.4 &  74.6 & \sota{77.0} & 87.0 & 70.8 & -8.2 \\
& QIG\vv{CVPR'26} 
& 39.9 & 91.9 & 34.3 & 69.6 & 67.7 & 72.6 & 74.7 & 86.5 & 67.2 & -12.8 \\

& \cc  
& 42.1 & {94.9} & \sota{55.3} & \sota{70.3} & \sota{71.1} & {75.0} & 76.9 & \sota{87.3} & \sota{71.6} & \sota{-7.1} \\
\hline

\rowcolor{lightgray}
% LLaVA-OV-7B & \textit{FP16}
% \multicolumn{2}{l}{LLaVA-OV-7B}
FP16 & LLaVA-OV-7B
& 46.0 &  95.8 & 60.4  & 76.1 & 62.2 &  80.2 & 81.3 & 90.2 &  74.0 & 100 \\
\hline
\multirow{4}{*}{\textit{W3A16}} &  AWQ\vv{MLSys'24}
& 43.1 & 94.5 & 60.0 & 73.2 & 59.8 & 76.8 & 78.3 & 86.2 & 71.5  & -3.4 \\
& MBQ\vv{CVPR'25} 
& \sota{45.0} & \sota{94.9} & 60.0 & 72.8 & 58.9 & \sota{77.7} &  79.1  & 86.3 & 71.8  & -3.0 \\
 & QIG\vv{CVPR'26} 
& 44.7 & 94.6 & \sota{61.2} & \sota{73.8} & 59.8 & 77.0 & 79.1 & 86.9 & 72.1  & -3.6 \\
 & \cc  
& 44.5 & 94.8  & 60.6 & 73.4 & \sota{60.3} & 77.4 & \sota{79.5} & \sota{86.9} & \sota{72.2}  & \sota{-2.4} \\
\hline
\multirow{4}{*}{\textit{W4A6}} & SQ\vv{ICML'23}  
& 39.8 & 90.9  & 54.3 & 55.9 & 36.9 & 65.0 & 74.1 & 64.6 & 60.2 & -18.6 \\
& MBQ\vv{CVPR'25}
& 40.0 & 90.1 & 55.9 & 64.0 & 49.3 & 72.3 & 74.1 & 79.5 & 65.6 & -11.4 \\
& QIG\vv{CVPR'26} 
& \sota{42.3} & 91.2 & 56.4 & 65.1 & \sota{50.1} & \sota{73.1} & 75.3 & 80.0 & 66.7 & -9.9 \\
& \cc
& 40.6 & \sota{91.6} & \sota{61.6} & \sota{66.0} & 49.7 & 72.7 & \sota{75.7} & \sota{81.4} & \sota{67.4} & \sota{-8.9} \\
\hline
\end{tabular}
}
\end{table*}

\begin{table}[!h]
\centering
\caption{\textbf{Quantization costs including gradient collection and (denoted as "+") CWS.} The standard (7B, 8B) and larger (26B, 32B) models are tested on one and four A100 cards, respectively.}
\label{tab:eff}
\resizebox{0.49\textwidth}{!}{
\begin{tabular}{llccc}
\toprule
\textbf{Setting} & \textbf{Models} &  \textbf{MBQ} & \textbf{\cc} & \textbf{QIG}  \\
\hline
\multirow{4}{*}{\textit{W3A16}}
& Qwen2.5VL-7B & 40s+10m42s  & 43s+10m42s & 41min15s \\
& InternVL2-8B & 40s+8m33s & 41s+8m47s & 27m0s \\
& Qwen2.5VL-32B & 2m37s+43m21s  & 2m50s+44m01s  & 89m38s  \\
& InternVL2-26B & 1m34s+20m34s  & 1m39s+21m01s & 59m35s  \\

\hline
\end{tabular}
}
\end{table}

\begin{table*}[h]
\centering
\caption{\textbf{Ablation study on the distribution of calibration set (Qwen2.5VL-7B).} We report results on different COCO subsets, sample sizes and Pile dataset (text-only). The default setting of 128 COCO samples is annotated by \colorbox{lightgray}{gray}.}
\label{tab:ablation_cal}
\resizebox{0.99\textwidth}{!}{
\begin{tabular}{llcccccccc|cc}
\toprule
\textbf{Setting} & \textbf{Method}  & \textbf{MMMU}  & \textbf{SQA} & \textbf{VizWiz} & \textbf{TextVQA} & \textbf{OCRBench} & \textbf{ChartQA}  & \textbf{AI2D} & \textbf{Doc} & \textbf{Avg} & \textbf{\%} \\
\hline
 \multirow{5}{*}{\textit{W3A16}} & COCO$_{64}$ 
& 48.2 & 85.0 & 64.4 & 79.5 & 82.4 & 80.7 & 79.7 & 94.7 & 76.8 & -3.6 \\
& \cellcolor{lightgray}COCO$_{128a}$ 
& \cellcolor{lightgray}48.4 & \cellcolor{lightgray}85.2 & \cellcolor{lightgray}64.6 & \cellcolor{lightgray}80.2 & \cellcolor{lightgray}82.6 & \cellcolor{lightgray}80.9 & \cellcolor{lightgray}79.9 & \cellcolor{lightgray}94.9 & \cellcolor{lightgray}77.1 & \cellcolor{lightgray}-3.3 \\
 & COCO$_{128b}$ 
& 48.2 & 85.6 & 64.0 & 79.9 & 82.8 & 80.7 & 80.1 & 94.7 & 77.0 & -3.4 \\
& COCO$_{256}$ 
& 48.9 & 85.3 & 65.0 & 79.5 & 82.2 & 80.4 & 79.9 & 94.9 & 77.0 & -3.4 \\

& Pile$_{128}$ 
& 47.8 & 85.5 & 66.0 & 80.6 & 81.0  & 79.1 & 80.7 & 94.7 & 76.9 & -3.5 \\
\hline
\multirow{4}{*}{\textit{W4A8}}  & COCO$_{64}$ 
& 45.4 & 84.7 & 65.6 & 77.0 & 77.2 & 81.3 & 79.2 & 90.1  & 75.1 & -5.8 \\
& \cellcolor{lightgray}COCO$_{128}$ 
& \cellcolor{lightgray}44.0 & \cellcolor{lightgray}85.0 & \cellcolor{lightgray}64.8 &\cellcolor{lightgray} 77.1 & \cellcolor{lightgray}77.2 & \cellcolor{lightgray}81.6 & \cellcolor{lightgray}78.9 &\cellcolor{lightgray} 90.0 & \cellcolor{lightgray}74.8 & \cellcolor{lightgray}-6.1 \\
& COCO$_{256}$ 
& 44.6 & 84.9 &  64.5 & 77.1 & 77.6 & 81.4 & 79.0 &  90.2  & 74.9 & -6.0 \\
 & Pile$_{128}$ 
& 44.3 & 82.2 & 63.5 & 75.6 & 81.0   & 79.1 & 80.7 & 88.0 & 74.3 & -6.8 \\

\hline
\end{tabular}
}
\end{table*}

\paragraph{Weight-activation quantization.}
As observed in Tab.~\ref{tab:results_qwen2.5}, MBQ \citep{li2025mbq} exhibits the worst performance on W4A8 and W4A6 scenarios of Qwen2.5VL. The token-wise QIG \citep{xiang2026qig} and hybrid MASQuant \citep{hu2026masquant} work better for weight-activation settings, while lagging behind ours by average scores. We attribute their unsatisfactory results to the misaligned importance and quantization, where our unified solution offers more benefits.
%%%%%%%%%%%%%%%%%%%%%%%%%%%
Meanwhile, we observe notable performance fluctuations of SQ \citep{xiao2023smoothquant} and MBQ \citep{li2025mbq} across different MLLMs in Tab.~\ref{tab:results_qwen2.5} and  Tab.~\ref{tab:results_intern2_llava}. For example, SQ outperforms MBQ on Qwen2.5VL-7B and -32B, whereas MBQ achieves a substantial advantage over SQ on LLaVA-OV-7B. One possible explanation is that the cascaded paradigm in early models like LLaVA-OV introduces stronger modality bias, which suits well with the motivation behind MBQ. In contrast, more advanced models such as Qwen2.5VL suffer less from these challenges. As a result, the gradient-based modality bias of MBQ becomes less reliable and even introduces additional noise, causing worse performance than the simpler SQ.
%%%%%%%%%%%%%%%%%%%%%%%%%%%%
In addition, the variance among datasets appears in low bitwidth settings. Considering the W4A6 on Qwen2.5VL-7B, SQ \citep{xiao2023smoothquant} gets merely 29.0 in MMMU, but yields the same accuracy score 92.4 as ours on DocVQA. 
In comparison, \cc~benefits from the unified channel-wise sensitivity and quantization to maintain SOTA performance across datasets and MLLM backbones.

\paragraph{Efficiency analysis.}
We analyze the efficiency from two aspects. First, as a channel-wise scaling approach, \cc~inherently offers superior quantization and inference efficiency compared to LoRA-based methods \citep{hu2026masquant,zhong2026breaking}, since a grid search is sufficient for the optimal channel-wise scales. Second, within the CWS family, the speed bottleneck is primarily determined by the reweighting strategy. 
As reported in Tab.~\ref{tab:eff}, \cc~achieves comparable quantization runtime with MBQ, while being substantially faster than QIG. 
To improve the binary importance metrics (from two modalities), QIG introduces a 32-step integration for sensitivity estimation.
Despite the fine-grained measure, it requires substantially longer time than MBQ. In comparison, we still maintain acceptable quantization costs.

\subsection{Ablation Study}
\label{subsec:exp_ablation}

\begin{figure}[!h]
\centering
\includegraphics[width=\linewidth]
{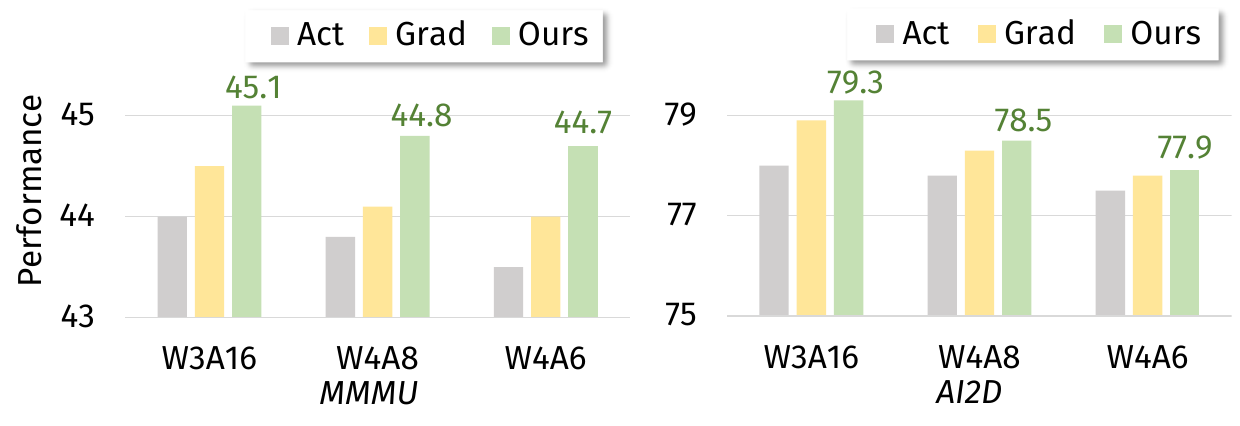} 
\caption{
\textbf{Ablation study on channel-wise importance (LLaVA-OV-7B).} We compare our diagonal Fisher against the mean absolute activation and gradient-based channel-wise importance metrics to highlight our significance.
}
\label{fig:ablation_ch}
\vspace{-0.5cm}
\end{figure}

\paragraph{How significant is our channel-wise metric?}
A natural hypothesis is that channel-wise sensitivity correlates with either the magnitude of activations or the magnitude of gradients. To examine this hypothesis and justify the use of Fisher Information, we replace the weighting factor $\mathbf{f}_j$ in Eqn.~(\ref{eq:err_c}) with two alternatives: the mean absolute activation values (\textit{i.e.}, $\mathrm{mean}_{:,j}(|\mathbf{y}_{ij}|)$) and the mean absolute gradient values (\textit{i.e.}, $\mathrm{mean}_{:,j}(|\mathbf{g}_{ij}|)$). 
The results in Fig.~\ref{fig:ablation_ch} validate the effectiveness of our Fisher-weighted objective over these heuristic designs across many settings.

\paragraph{How stable is the empirical Fisher?}
To evaluate the robustness of our method, we conduct ablation study on $\mathcal{D}_\mathrm{cal}$. 
As shown in Tab.~\ref{tab:ablation_cal}, two randomly sampled COCO subsets (COCO$_{128a},_{128b}$) yield nearly identical results.
This demonstrates that \cc~is insensitive to the choice of specific calibration samples. Furthermore, when the calibration size is adjusted to 64 or 256 samples, the average performance fluctuates within $\pm 0.3$\%, confirming the method's robustness to sample size variations. Finally, while calibrating on text-only data from Pile causes a minor performance drop of 0.2$\sim$0.5, \cc~flexibly supports both multimodal and text-only calibration. In contrast, modality-aware approaches like MBQ are strictly limited to multimodal data.

\begin{table*}[!h]
\centering
\caption{\textbf{Comparison with GPTQ and RTN (W3A16).} GPTQ shows even worse performance then the naive round-to-nearest (RTN), reflecting the significance of alignment between quantization error and loss perturbation.}
\label{tab:ablation_gptq}
\resizebox{0.99\textwidth}{!}{
\begin{tabular}{llcccccccc|cc}
\toprule
\textbf{Model} & \textbf{Method}  & \textbf{MMMU}  & \textbf{SQA} & \textbf{VizWiz} & \textbf{TextVQA} & \textbf{OCRBench} & \textbf{ChartQA}  & \textbf{AI2D} & \textbf{Doc} & \textbf{Avg} & \textbf{\%} \\
\hline
\multirow{3}{*}{{Qwen2.5VL-7B}}
& RTN 
& 48.0 & 83.6 & 67.4 & 79.1 & 78.8 & 75.0 & 79.0 & 92.5 & 75.4 & -5.4 \\
& GPTQ\vv{ICLR'23}
& 48.2 & 73.4 & 65.5 & 78.7 & 76.4 & 62.5 & 79.2 &  87.8 & 71.5 & -10.3 \\
& \cc
& 48.4 & 85.2 & 64.6 & 80.2 & 82.6 & 80.9 & 79.9 & 94.9 & \sota{77.1} & \sota{-3.3} \\
\hline
 \multirow{3}{*}{{InternVL2-8B}} & RTN 
& 43.8  & 96.2 & 56.1 & 74.6 & 73.7 & 79.4 & 80.5 & 90.2 & 74.3 & -3.6 \\
& GPTQ\vv{ICLR'23}
& 43.1 & 94.3 & 59.8 & 74.5 & 74.1 & 76.4 & 76.6 & 90.2 & 73.6 & -4.5 \\
 & \cc
& 45.5 & 96.5 & 59.9 & 75.0 & 75.9 & 80.4 & 80.4 & 90.5  & \sota{75.5} & \sota{-2.1} \\

\hline
\end{tabular}
}
\end{table*}

\begin{table}[!h]
\centering
\caption{\textbf{Ablation study on dense Fisher and first-order derivatives (Qwen2.5VL-7B).} We report average performance over four benchmarks and the quantization time in the last two columns.}
\label{tab:ablation_dense}
\resizebox{0.49\textwidth}{!}{
\begin{tabular}{lcccc|cc}
\toprule
\textbf{Method} & \textbf{MMMU}  & \textbf{SQA} & \textbf{OCRBench} & \textbf{ChartQA}  &  \textbf{Avg} & \textbf{Time} \\
\hline
\rowcolor{lightgray}
\multicolumn{7}{c}{\textit{W3A16}} 
\\
% \hline
\rowcolor{white}
Dense
& 47.7  & 84.2  & 82.2 & 79.9 & 73.5 & 50 min\\
+1.order  
& 48.3 &  85.3  & 82.4 & 80.5 & 74.1 & 12 min\\
Ours  
& 48.4 &  85.2  & 82.6 & 80.9 & 74.3 & 11 min\\
\hline
\rowcolor{lightgray}
\multicolumn{7}{c}{\textit{W4A8}} 
\\
% \hline
\rowcolor{white}
Dense &
44.8 &  84.7  & 77.3  & 81.7 & 72.1 & 62 min\\
+1.order 
& 44.4 & 85.0  & 77.0 & 81.5 & 72.0 & 19 min \\
Ours  
& 44.0 & 85.0  & 77.2 & 81.6 & 72.0 & 18 min \\
\hline
\end{tabular}
}
\end{table}

\paragraph{Does the diagonal approximation remain faithful to the original dense Fisher matrix?}
We validate the effectiveness of our diagonal approximation by comparing it directly against the full dense Fisher matrix. Tab.~\ref{tab:ablation_dense} demonstrates that the diagonal simplification successfully captures important channels with comparable or even higher performance. Meanwhile, our diagonal Fisher achieves substantial savings in computational resources, as shown in the quantization time.

\paragraph{Can the first-order term be safely ignored?}
In line with OBD \citep{lecun1989optimal}, we omit the first-order term in the loss perturbation expansion to minimize computational costs. As demonstrated in Tab.~\ref{tab:ablation_dense}, incorporating this term introduces significant overhead with remains marginal difference, thereby validating our objective in Eqn.~(\ref{eq:delta_l_c}). Further details are provided in App.~\ref{subsec:app_ablation}.

\paragraph{Comparison with RTN and GPTQ.} As a Hessian-motivated method, we compare \cc~ with GPTQ \citep{frantar2023gptq} and round-to-nearest (RTN) in Tab.~\ref{tab:ablation_gptq}.
Consistent with previous observations \citep{li2025mbq,deng2026vlm}, GPTQ performs even worse than the naive RTN on the latest MLLMs.
%%%%%%%%%%%%%%%%%%%%%%%%
This degradation suggests that solely minimizing layer-wise quantization error does not guarantee optimal performance, as the activation-derived Hessian in GPTQ is not explicitly aligned with the SFT objective. Ultimately, this highlights the effectiveness of our framework, which directly aligns the quantization error with the task-specific loss landscape rather than merely relying on reconstruction error.

%%%%%%%%%%%%%%%%%%%%%%%%%%%%%%%%%%

\section{Conclusion}
\label{sec:conclusion}

In this paper, we propose \cc, a unified channel-wise post-training quantization method that explicitly connects task-specific loss perturbation with quantization error. To align the sensitivity metric and quantization error at the channel level, 
% we derive a Fisher-weighted objective by employing a diagonal Fisher matrix as a tractable Hessian approximation.
we employ a diagonal Fisher matrix from SFT loss as a tractable Hessian approximation to measure the channel-wise sensitivity.
Based on that, we propose a Fisher-weighted objective,  
selectively penalizing errors in loss-critical channels while significantly reducing the computational overhead compared to dense matrices. Operating as a lightweight framework, \cc~achieves SOTA performance with exceptional efficiency. Extensive experiments on Qwen2.5-VL, InternVL2, and LLaVA-OV across eight benchmarks demonstrate the effectiveness of our approach in both weight-only and weight-activation quantization regimes.

\noindent \textbf{Limitations and future work.} Despite the effectiveness of our unified framework, our current quantization scope is limited to LLM decoders, leaving the vision encoder and projection layers in full precision. Furthermore, similar to existing CWS approaches, \cc~experiences severe performance degradation under extreme compression settings (\textit{e.g.}, W4A4). Extra compensation is necessary for extremely low-bit scenarios, which is considered as our future work.

% \section{Copyright}
% .....

\section{Acknowledgments}

This research received funding from the Flemish Government (AI Research Program) and the Research Foundation
Flanders (FWO) through project number G0G2921N.
We acknowledge EuroHPC JU for awarding access to LEONARDO hosted by CINECA (Italy) and the LEONARDO consortium.

\bibliography{main}

\clearpage

\appendix

\twocolumn[
\begin{center}
    \LARGE\bfseries Appendix
    \vspace{0.4cm}
\end{center}
]

\section{Extended Experiments}
\label{sec:app_exp}

\subsection{Implementation Details}
\label{subsec:app_imple}

The smaller models ($\leq$8B) are evaluated on a single A100 GPU, and larger models (26B$\sim$32B) are deployed on 4 A100 GPUs. We summarize the LLM backbone and number of layers for MLLMs used in the main paper in Tab.~\ref{tab: app_llm}. To ensure a fair comparison, all PTQ baselines are evaluated using the same subset of the COCO dataset, sampled with a fixed seed of 42. We implement FlashAttention \citep{dao2023flashattention2} for QwenVL- and InternVL-series, and adopt eager attention for the LLaVA-OV model.
Due to the vast numerical discrepancy between gradients and MSE, we normalize the Fisher factors by their mean value, written as
$\mathbf{f}_j{=}{\mathbf{f}_j}/{\bar{\mathbf{f}}}$. Experiments validate that the normalization choice doesn't affect the performance, \textit{e.g.}, mean, MinMax an so on.

\begin{table}[h]
\centering
\caption{Overview of MLLMs.}
\label{tab: app_llm}
\setlength{\tabcolsep}{5pt} 
\begin{tabular}{l lc} 
\hline
MLLM & LLM & \# Layers \\
\hline
Qwen2VL-2B & Qwen2-2B  &  28  \\
Qwen2VL-7B & Qwen2-7B  &  28  \\
Qwen2.5VL-7B & Qwen2.5-7B  &  28  \\
Qwen2.5VL-32B & Qwen2.5-32B  &  64   \\
\hline
InternVL2-8B & InternLM2.5-7B  &  32 \\
InternVL2-26B & InternLM2-20B  & 48  \\
\hline
LLaVA-OV-7B & Qwen2-7B  &  28   \\

\hline
\end{tabular}
\end{table}

\begin{figure*}[h]
\centering
\includegraphics[width=\linewidth]
{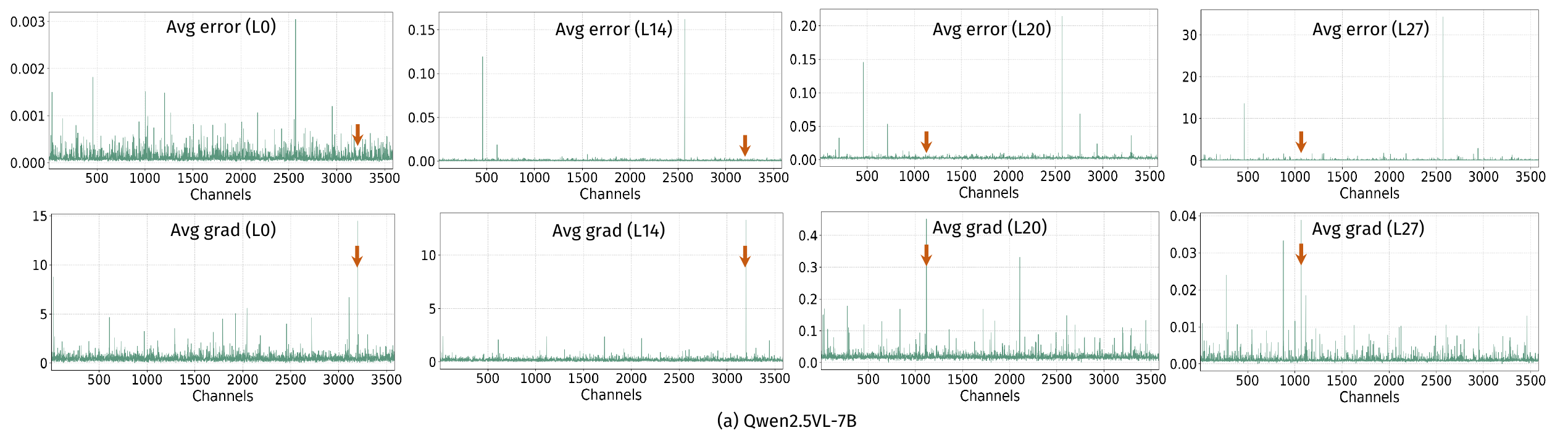} 
\includegraphics[width=\linewidth]
{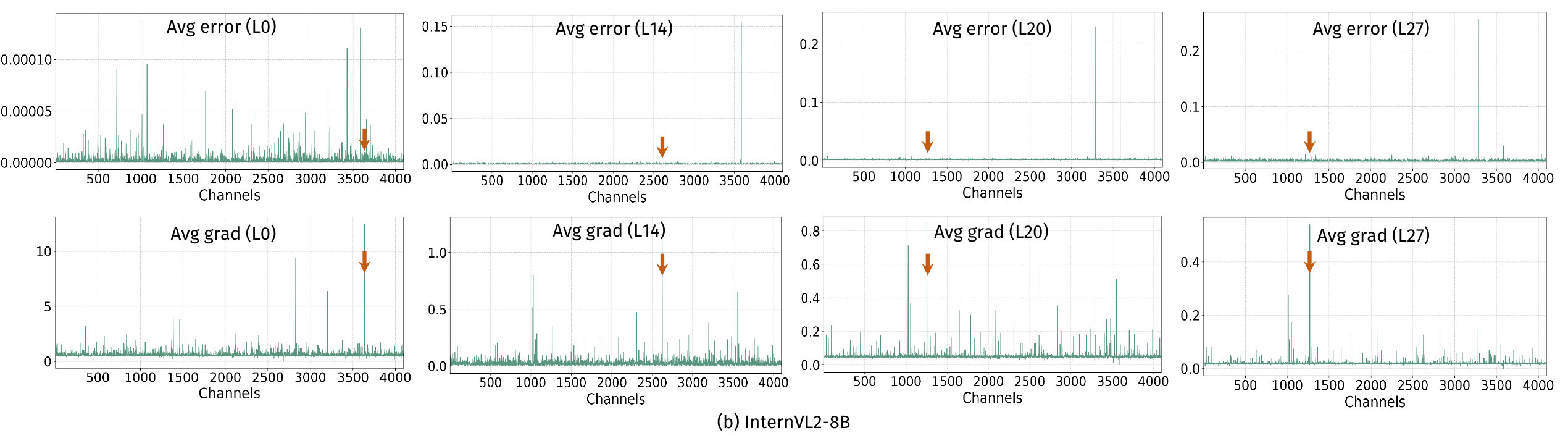} 
\caption{
\textbf{Visualization of channel-wise quantization error and gradients (Qwen2.5VL-7B, InternVL2-7B).} We recommend viewing the layer-wise subfigure vertically. The red arrows indicate misaligned channels, where small errors correspond to disproportionately large gradients. Under such misalignment, optimizing towards minimizing quantization error can induce excessively large gradients and poor downstream performance, which motivates our channel-wise aligned method.
}
\label{fig:vis_ch_long}
\end{figure*}

\subsection{Dataset Description}
\label{subsec:app_dataset}

\textbf{MMMU} \citep{yue2023mmmu}. 
MMMU was designed to evaluate multimodal models on massive multi-discipline tasks demanding college-level subject knowledge and deliberate reasoning. The questions in MMMU span 30 subjects and 183 subfields. Meanwhile, it comprises 32 image types, such as charts, diagrams, maps, tables, music sheets and chemical structures. MMMU highlights advanced perception and reasoning with domain-specific knowledge, which demands that MLLMs tackle tasks requiring expert-level proficiency.
%%%%%%%%%%%%%%%%%%%%%%%%
The validation set used in our experiments has 
% \colorbox{beige}{900}
\textbf{900}
multiple-choice question-answer pairs.

\textbf{SQA-IMG} \citep{lu2022learn}.
We denote SQA for ScienceQA benchmark, which employs multiple-choice questions to assess a model's performance in the scientific domains. The dataset spans three primary subjects, \textit{i.e.}, natural science, language science and social science. It features a hierarchical structure organized by topic, category and skill. This hierarchy comprises 26 topics, 127 categories and 379 skills. While the questions are accompanied by relevant illustrations, a portion of the dataset is text-only. 
%%%%%%%%%%%%%%%
We evaluate on the SQA-IMG subset, which consists of
% \colorbox{beige}{2,017}
\textbf{2,017}
multimodal pairs where both images and questions are present.
%%%%%%%%%%%%%%%

\textbf{VizWiz} \citep{gurari2018vizwiz}.
VizWiz is a real-world visual question answering dataset designed to assist people who are blind or visually impaired. Unlike traditional VQA datasets that are collected from carefully curated images, VizWiz consists of images captured by visually impaired users using their mobile devices and the corresponding spoken questions they asked. As a result, the dataset contains many real-world challenges, including poor image quality, occlusions, blur, improper framing and unanswerable questions. These characteristics make it a more realistic benchmark for evaluating the robustness and practical applicability of VQA systems.
%%%%%%%%%%%%%%%%%%%%%%%%
In this paper, we test on the validation set with 
% \colorbox{beige}{4,319}
\textbf{4,319}
questions.

\textbf{OCRBench} \citep{liu2024ocrbench}.
OCRBench is a comprehensive benchmark designed to evaluate the Optical Character Recognition (OCR) capabilities of MLLMs. It contains question-answer pairs collected from 29 publicly available datasets and covers five representative OCR-related tasks: text recognition, scene text-centric VQA, document-oriented VQA, key information extraction and handwritten mathematical expression recognition. While traditional OCR benchmarks focus solely on text transcription accuracy, OCRBench provides a unified evaluation framework for assessing both text perception and reasoning abilities in multimodal models. 
%%%%%%%%%%%%%%%%%%%%%%%%
OCRBench contains 
% \colorbox{beige}{1,000}
\textbf{1,000}
questions.

\textbf{TextVQA} \citep{singh2019towards}.
The TextVQA benchmark aims to assess a model’s proficiency in reading and reasoning over visual text. It emphasizes the integration of Optical Character Recognition (OCR) with natural language understanding. The images, primarily sourced from OpenImages-v3 \citep{openimages}, feature diverse real-world scenarios—such as street signs, billboards and product packaging—that are rich in textual content. Alongside the raw images, ground-truth OCR tokens are provided as auxiliary input. To arrive at the correct answer, models must either extract text directly from the image or perform contextual reasoning based on the identified characters. 
%%%%%%%%%%%%%%%%%%%%%%%
We report evaluation results on the validation set, which comprises 
% \colorbox{beige}{5,000}
\textbf{5,000}
image-question pairs.

\textbf{ChartQA} \citep{masry2022chartqa}.
ChartQA is a dataset for chart question answering that evaluates a model’s ability to understand and reason over information presented in charts. ChartQA emphasizes both visual understanding and complex logical reasoning, requiring models to perform operations such as comparison, aggregation, arithmetic calculation and trend analysis based on chart content. The benchmark covers various chart types, including bar charts, line charts and pie charts, and often requires integrating visual cues with underlying numerical information. 
%%%%%%%%%%%%%%%%%%%%%%%%
Overall, it provides 
% \colorbox{beige}{2,500}
\textbf{2,500}
questions for evaluation.

\textbf{AI2D} \citep{kembhavi2016diagram}.
AI2D focuses on diagrams, which are common tools for representing complex concepts, relationships and events. The challenge lies in identifying the diagram structure, the constituent semantics and their relationships. 
Subsequently, it provides detailed ground truth for diagram parsing, including the segmentation of visual elements like arrows, text boxes and blobs, as well as their syntactic relations.
%%%%%%%%%%%%%%%%%%%%%%%%
AI2D consists of 
% \colorbox{beige}{3,088}
\textbf{3,088}
questions.

\textbf{DocVQA} \citep{mathew2021docvqa}.
DocVQA is a large-scale dataset specifically designed for VQA tasks on document images, including thousands of pages from diverse real-world documents.
It challenges AI models to understand both text and layout structure by requiring them to answer natural language questions based on invoices, reports, forms and scanned scripts.
%%%%%%%%%%%%%%%%%%%%%%%%
The validation set has 
% \colorbox{beige}{5,349}
\textbf{5,349}
questions.

\subsection{More Results}
\label{subsec:app_ablation}

Here, we extend the discussion from Sec.~\ref{subsec:exp_ablation}. First, we demonstrate that \cc~is robust across various model scales ranging from 2B to 72B. Next, we provide the derivation incorporating the first-order term reported in Tab.~\ref{tab:ablation_dense}, along with its application to LLM quantization. Finally, we present additional visualizations illustrating the misalignment between the quantization error and SFT gradients on Qwen2.5VL and InternVL2.

\begin{figure}[!h]
\centering
\includegraphics[width=\linewidth]
{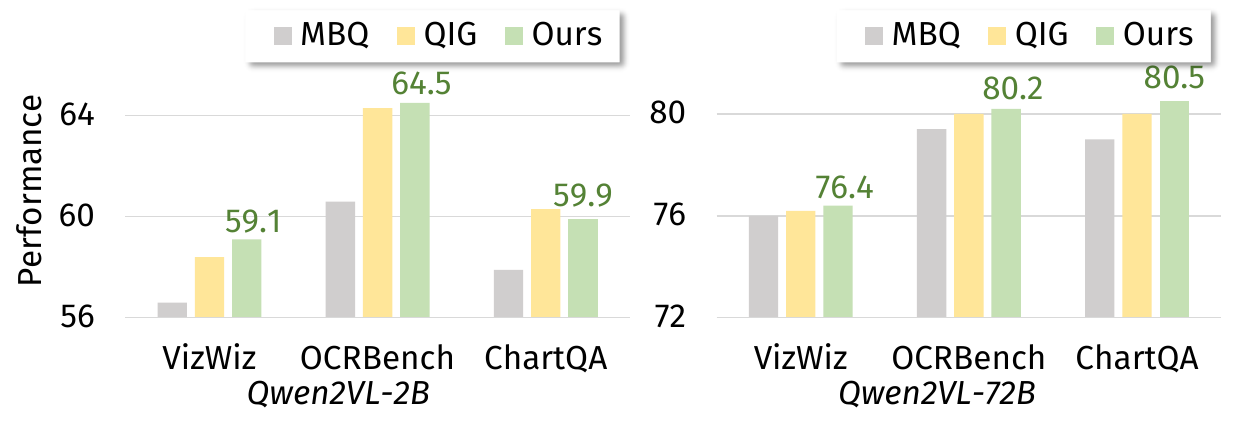} 
\caption{
\textbf{Ablation study on more model scales (W3A16).} Our method maintains SOTA performance across scales.
}
\label{fig:ablation_qwen2}
\end{figure}

\paragraph{Does \cc~maintain SOTA performance across varying model scales?}
Beyond the 7B to 32B models evaluated in Sec.~\ref{subsec:exp_results}, we further quantize the 2B and 72B versions of Qwen2VL. Fig.~\ref{fig:ablation_qwen2} compares our approach with MBQ and QIG under the W3A16 setting across three datasets. The consistent improvements indicate that our unified channel-wise scaling framework generalizes highly well to both small (2B) and large (72B) models.

\paragraph{Can the first-order term be safely ignored?}
If we consider both first and second order in Eqn.~(\ref{eq:delta_l}), then the loss perturbation $\Delta\mathcal{L}$ in Eqn.~(\ref{eq:delta_l_c}) is formulated as:
\begin{align}
    \Delta\mathcal{L}\ &\approx
    \mathbf{g}^{\top} \Delta\mathbf{Y}+\frac{1}{2}\Delta\mathbf{Y}^{\top}\mathrm{diag}(\mathbf{f})\,\Delta\mathbf{Y}\
    \label{eq:delta_l_d}
    \\
    &= \sum_{i=1}^{N_k} (\mathbf{g}^{\top} \Delta\mathbf{y}_i+\frac{1}{2} \|\mathbf{\sqrt{f}} \odot
\Delta\mathbf{y}_i \|^2).
\label{eq:delta_l_e}
\end{align}
Leveraging the triangle inequality, we obtain the following upper bound on the loss perturbation:
\begin{align}
    \Delta\mathcal{L}&\leq \sum_{i=1}^{N_k} (|\mathbf{g}|^{\top} |\Delta\mathbf{y}_i|+\frac{1}{2} \|\mathbf{\sqrt{f}} \odot
\Delta\mathbf{y}_i \|^2).
    \label{eq:delta_l_first_order}
\end{align}
The experimental results are reported in Tab.~\ref{tab:ablation_dense}. Since retaining the first-order term yields negligible benefits yet introducing substantial computational overhead, our elegant formulation in Eqn.~{(\ref{eq:delta_l_c})} is justified.

\paragraph{Applications in LLMs.}
Our method is also readily applicable to LLMs such as Llama2-7B \citep{touvron2023llama2} and Llama3-8B \citep{grattafiori2024llama3}. Following AWQ \citep{lin2023awq}, we quantize LLMs using 128 calibration samples from the Pile dataset under the W4A16 setting. Tab.~\ref{tab:ablation_llm} reports the evaluation across four standard benchmarks: perplexity (PPL), PIQA \citep{bisk2020piqa} and the easy and challenge subsets of ARC \citep{clark2018think}. Specifically, we utilize the validation set of PIQA (3,676 samples) and the test sets of ARC-easy (9,496 samples) and ARC-challenge (4,687 samples). The average scores across PIQA, ARC-easy, and ARC-challenge indicate that \cc~consistently outperforms AWQ. Ultimately, this successful extension demonstrates the broad applicability and effectiveness of our framework.

\begin{table}[!h]
\centering
\caption{\textbf{Performance on LLM quantization (W4A16).} }
\label{tab:ablation_llm}
\resizebox{0.46\textwidth}{!}{
\begin{tabular}{lccccc}
\toprule
\textbf{Method} & \textbf{PPL}$_\downarrow$  &\textbf{PIQA}$_\uparrow$ &\textbf{ARC-easy}$_\uparrow$ & \textbf{ARC-chal.}$_\uparrow$ &  \textbf{Avg}$_\uparrow$ \\
\hline
\rowcolor{lightgray}
\multicolumn{6}{c}{\textit{Llama2-7B}} 
\\
\rowcolor{white}
AWQ
& 5.59  & 78.3  & 68.9 & 41.1 & 62.8 \\
\cc
& 5.60  & 78.5  & 69.6 & 41.5 & 63.2 \\
\hline
\rowcolor{lightgray}
\multicolumn{6}{c}{\textit{Llama3-8B}} 
\\
\rowcolor{white}
AWQ
& 6.53  & 78.9  & 79.6 & 49.9 & 69.5 \\
\cc
& 6.49  & 79.4  & 80.1 & 50.3 & 69.9 \\
\hline

\end{tabular}
}
\end{table}

\paragraph{Visualization of the misaligned quantization error and SFT gradients.}
As shown in Fig.~\ref{fig:vis_ch_long}, this misalignment is evident across multiple layers in both Qwen2.5VL and InternVL2. Specifically, the red arrows in the top panel of Fig.~\ref{fig:vis_ch_long}(a) and (b) highlight channels with small quantization errors, which vanilla AWQ \citep{lin2023awq} treats as an indicator of low sensitivity. However, we observe that these corresponding channels actually exhibit large SFT gradients. This discrepancy demonstrates that the optimization towards minimizing local quantization error does not necessarily preserve downstream performance. Ultimately, this visualization corroborates the necessity of the quadratic formulation in Eqn.~(\ref{eq:second_order}).

\section{Extended Related Work}
\label{sec:app_related}

\paragraph{Fisher Information for channel-wise sensitivity.}
While Fisher Information is widely adopted in model compression \citep{liu2021group,xu2025towards}, its application varies significantly. 
For instance, Supernode \citep{cherilyn2026supernodes} leverages a Fisher-based metric to assess the loss perturbation of masking a channel, and defines a "supernode" from the Fisher-style loss proxy (LP) for channel pruning. Since the top 1\% channels dominate the overall LP mass, they protect these "supernodes" from being pruned under a given sparsity budget. In contrast, we employ Fisher as a weighting factor for post-training quantization.
Meanwhile, Supernode tackles the intertwined expectation of the joint second moments of activation and gradient in a computationally expensive manner. As described in Sec.~\ref{subsec:method_Fisher}, we adopt a diagonal approximation for efficient measurement of channel-wise sensitivity.
Another relevant work, High-Impact-Ratio \citep{pham2026layer} proposes a mixed-precision weight quantization for LLMs, optimizing the layer-wise ratio of high-impact parameters instead of a fixed ratio across layers.
The average Fisher Information serves as an impact score to determine the optimal ratio of high-impact parameters per layer under a fixed resource budget. Later, the selected high-impact parameters are quantized with a higher bit-width and a stronger quantization method, and vice versa. Unlike such bit-width scheduling methods \cite{pham2026layer}, \cc~is a principled PTQ framework designed to fix the objective misalignment inherent in channel-wise scaling. Furthermore, mixed-precision quantization often brings challenges for hardware deployment due to heterogeneous bit-width execution, whereas \cc~offers a hardware-friendly quantization scheme to avoid such issues.

\end{document}